\documentclass{bmvc2k}

\title{IronDepth: Iterative Refinement of Single-View Depth using Surface Normal and its Uncertainty}

\addauthor{Gwangbin Bae}{gb585@cam.ac.uk}{1}
\addauthor{Ignas Budvytis}{ib255@cam.ac.uk}{1}
\addauthor{Roberto Cipolla}{rc10001@cam.ac.uk}{1}

\addinstitution{
Department of Engineering\\
University of Cambridge\\
Cambridge, UK
}

\runninghead{Bae, Budvytis, Cipolla}{IronDepth}

\usepackage{multirow}
\usepackage{array}
\usepackage{booktabs}
\usepackage{makecell}
\usepackage{mathtools}

\usepackage{tikz}
\usepackage{comment}
\usepackage{amsmath,amssymb} 
\usepackage{color}

\begin{document}

\maketitle

\begin{abstract}
Single image surface normal estimation and depth estimation are closely related problems as the former can be calculated from the latter. However, the surface normals computed from the output of depth estimation methods are significantly less accurate than the surface normals directly estimated by networks. To reduce such discrepancy, we introduce a novel framework that uses surface normal and its uncertainty to recurrently refine the predicted depth-map. The depth of each pixel can be propagated to a query pixel, using the predicted surface normal as guidance. We thus formulate depth refinement as a classification of choosing the neighboring pixel to propagate from. Then, by propagating to sub-pixel points, we upsample the refined, low-resolution output. The proposed method shows state-of-the-art performance on NYUv2~\cite{NYUv2} and iBims-1~\cite{iBims} - both in terms of depth and normal. Our refinement module can also be attached to the existing depth estimation methods to improve their accuracy. We also show that our framework, only trained for depth estimation, can also be used for depth completion. The code is available at \url{https://github.com/baegwangbin/IronDepth}.
\end{abstract}


\begin{figure}[t]
\begin{center}
\includegraphics[width=1.0\linewidth]{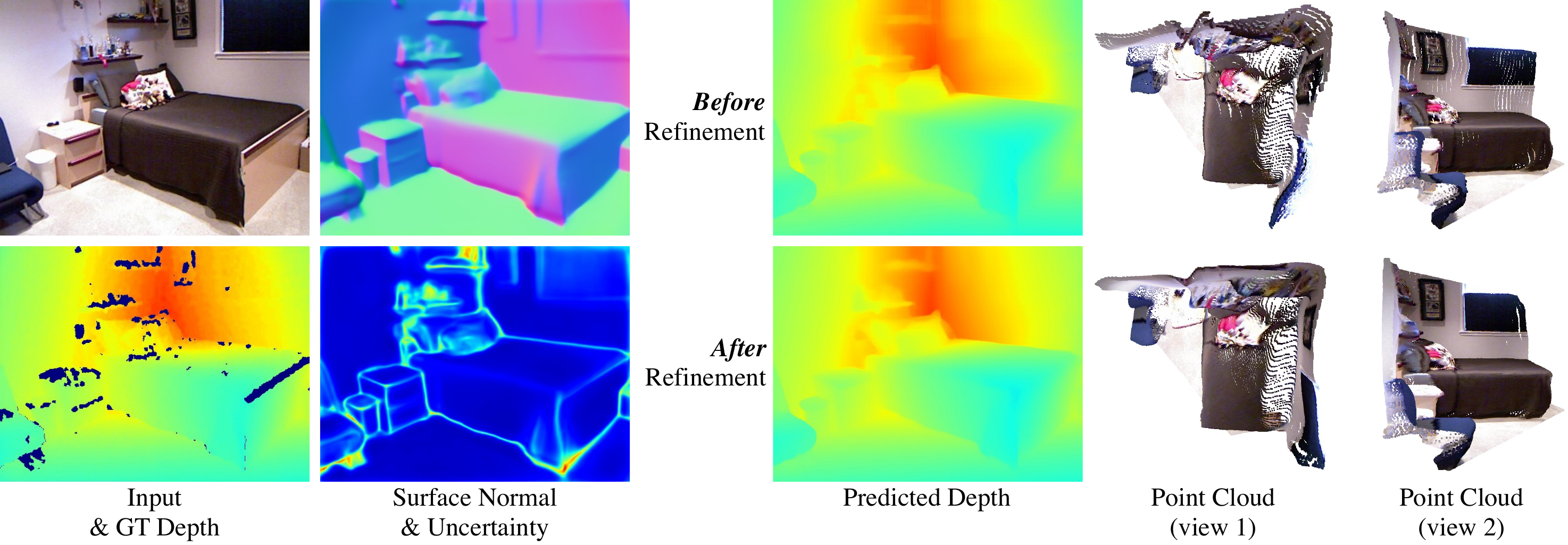}
\end{center}
\caption{This figure shows how the proposed normal-guided depth refinement improves the quality of the 3D reconstruction. While the predicted depth-maps look similar, the point cloud comparison shows that the characteristics in the scene geometry (e.g., orientation of the walls) are better preserved in the refined output.}
\label{fig:intro}
\end{figure}

\section{Introduction}
\label{sec:intro}

Monocular 3D reconstruction is one of the fundamental problems in computer vision, with a wide variety of applications including autonomous driving~\cite{KITTI}, augmented reality~\cite{SNfromRGB_19_FrameNet}, and 3D photography~\cite{3d_photography}. In this paper, we focus on two popular approaches in single image 3D reconstruction - surface normal estimation and depth estimation. The two problems are closely related as surface normal can also be computed from the predicted depth-map.

For both tasks, deep learning-based methods have shown impressive performance. However, if we calculate the surface normal from the output of depth estimation methods, its accuracy is significantly worse than that of surface normal estimation methods. For NYUv2~\cite{NYUv2} dataset, the surface normal calculated from the depth-map predicted by AdaBins~\cite{mono-2020-adabins} has mean angular error of $28.8^\circ$, which is nearly twice as large as $14.9^\circ$ achieved by the direct estimation of Bae et al.~\cite{SNfromRGB_21_BAE}. This suggests that, while depth estimation methods show low \textit{per-pixel} depth errors, the recovered \textit{surface} does not faithfully capture the characteristics of the scene geometry (e.g., the walls and floors are not flat).

\begin{figure}[t]
\begin{center}
\includegraphics[width=1.0\linewidth]{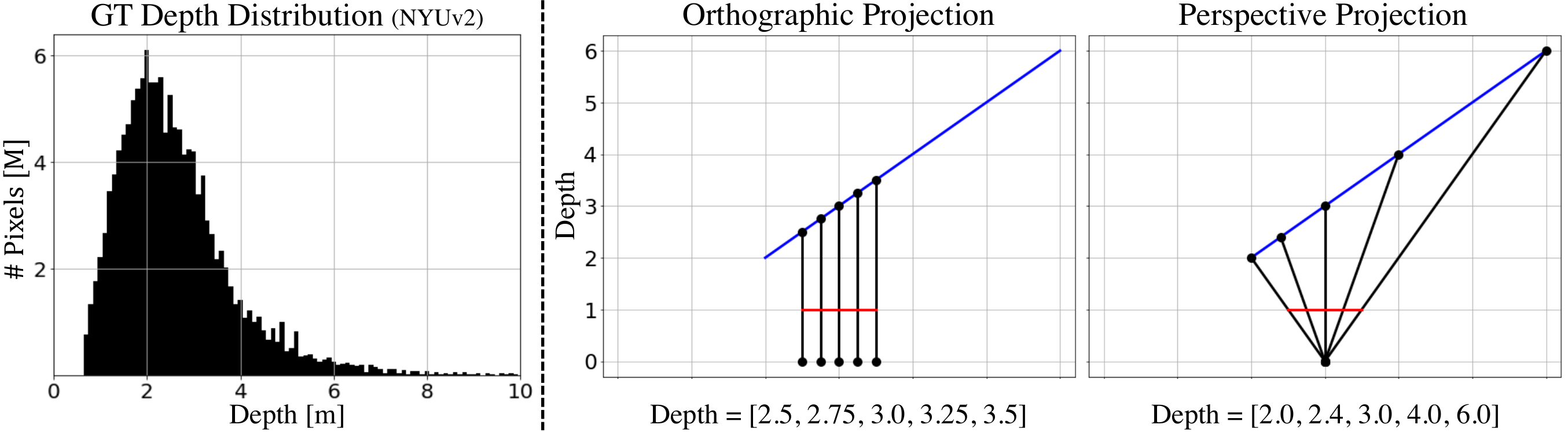}
\end{center}
\caption{\textbf{(left)} This histogram shows the distribution of the ground truth depth in NYUv2~\cite{NYUv2} test set. Most of the pixels have depth of 1-4m, making the prediction biased to those intermediate values. \textbf{(right)} In this figure, the blue, red and black lines represent a flat surface, image plane and pixel rays, respectively. For orthographic projection, the depth values are linear within the image plane, and the gradient of depth is constant. For perspective projection, however, the depth gradient is dependent on the viewing direction, making the inference challenging for CNNs, which are designed to be translation-equivariant.}
\label{fig:motivation}
\end{figure}

Poor surface normal accuracy of depth estimation methods is mainly caused by two problems. First is the \textit{imbalance in the training data}. Fig.~\ref{fig:motivation}-(left) shows that most of the pixels in NYUv2~\cite{NYUv2} have ground truth depth of 1-4m. If depth estimation is solved as \textit{regression}, the network is biased to predict those intermediate depth values, leading to poor surface normal accuracy. Secondly, estimating a depth-map with high surface normal accuracy requires \textit{view-dependent} inference. Fig.~\ref{fig:motivation}-(right) shows that, for perspective camera, the depth-map corresponding to a flat surface is not linear. Unlike surface normal, the depth gradient is not constant within the surface and is dependent on the viewing direction. Depth estimation is thus difficult to solve using convolutional neural networks, which are designed to be translation-equivariant.

In this paper, we propose IronDepth, a novel framework that uses surface normal and its uncertainty to recurrently refine the initial depth-map (\textbf{iron}: \textbf{i}terative \textbf{r}efinement \textbf{o}f depth using \textbf{n}ormal). Given the estimated depth and surface normal of a pixel, we can define a plane. Then, for a query pixel, we can calculate how its depth should be updated in order for it to belong to the same plane. We call this the \textit{normal-guided depth propagation}. We then formulate depth refinement as \textit{classification} of choosing the neighboring pixel to propagate from. After refining the initial depth-map in a coarse resolution, we apply the same normal-guided depth propagation to sub-pixel points to upsample the refined output. Fig.~\ref{fig:intro} shows how the proposed normal-guided refinement improves the quality of the 3D reconstruction.



Our method achieves state-of-the-art performance on NYUv2~\cite{NYUv2}. We also outperform other methods in cross-dataset evaluation on iBims-1~\cite{iBims}. While the improvement in depth accuracy is small, the surface normal calculated from our depth prediction is significantly more accurate than those obtained by the competing methods.

We also run additional experiments to further investigate the usefulness of the proposed surface normal-guided depth refinement module. Firstly, the initial depth prediction can be replaced with the output of the existing depth estimation methods to improve their accuracy. We confirmed this by applying our framework to five state-of-the-art depth estimation methods. Secondly, our framework can seamlessly be applied to depth completion. Given a sparse depth measurement, we can fix the depth for the pixels with measurement. This allows the information (i.e. sparse depth measurements) to be propagated to the neighboring pixels, improving the overall accuracy.

\section{Related Work}

\noindent
\textbf{Single image depth estimation.} The goal of the problem is to estimate the per-pixel metric depth. Owing to the advances in deep neural networks (DNNs), state-of-the-art approaches~\cite{mono-2014-eigen,mono-2015-eigen,mono-2015-liu,mono-2016-laina,mono-2017-cao,mono-2017-kuznietsov,mono-2018-DORN,mono-2019-BTS,mono-2019-VNL,mono-2020-adabins,mono-2020-dav,mono-2021-transdepth} use DNNs to extract features and predict the per-pixel metric depth. While most methods solve depth estimation as regression, other methods recast the problem as classification~\cite{mono-2017-cao,mono-2020-adabins} or ordinal regression (i.e. classification on ordered thresholds)~\cite{mono-2018-DORN} by discretizing the output depth. We propose a hybrid approach where the initial depth-map is obtained via regression and then refined by solving classification of selecting the neighboring pixel to propagate from. 

\noindent
\textbf{Single image surface normal estimation.} The goal of the problem is to estimate the per-pixel surface normal vector, defined in the camera-centered coordinates. Similar to depth estimation, this problem is solved via direct regression using DNNs~\cite{SNfromRGB_15_Deep3D,SNfromRGB_15_Eigen,SNfromRGB_16_SkipNet,SNfromRGB_19_SR,SNfromRGB_20_TiltedSN,SNfromRGB_21_BAE}. Notable contributions have been made by using a spatial rectifier to improve the performance on tilted images~\cite{SNfromRGB_20_TiltedSN}, and using vision transformers~\cite{other-ViT1,other-ViT2} to encode the global context~\cite{mono-2021-transdepth}. While most methods only estimate the normal, recent work by Bae et al.~\cite{SNfromRGB_21_BAE} also estimates the associated uncertainty. They also proposed to apply the training loss on a subset of pixels selected based on the estimated uncertainty, thereby improving the quality of prediction on small structures and near object boundaries. 

\noindent
\textbf{Improving depth estimation using surface normal.} Many attempts have been made to exploit the relationship between depth and surface normal. Yin et al.~\cite{mono-2019-VNL} proposed virtual normal loss, where triplets of pixels are sampled during training and the surface normal of the triangle is computed from the predicted depth and the ground truth. They applied L1 loss between the computed normals. Long et al.~\cite{mono-2021-ASN} improved upon this work by adaptively combining the normals computed for different triplets. Our normal-guided depth propagation is inspired by GeoNet++~\cite{SNfromRGB_20_GeoNet++}, which iterates between depth-to-normal and normal-to-depth modules. The difference is three-fold. Firstly, the normal-to-depth module in~\cite{SNfromRGB_20_GeoNet++} is deterministic (i.e. no learnable parameter). The propagation weight between pixel $i$ and $j$ is determined by their surface normal similarity, $\mathbf{n}_i^\intercal \mathbf{n}_j$. This can fail if $i$ and $j$ belong to disconnected planes with similar surface normals. Instead, we learn the propagation weights in a recurrent framework. Secondly, we use surface normal uncertainty to avoid propagating from the pixels with high uncertainty. Lastly, we extend the normal-guided depth propagation to depth upsampling.


\begin{figure}[t]
\begin{center}
\includegraphics[width=1.0\linewidth]{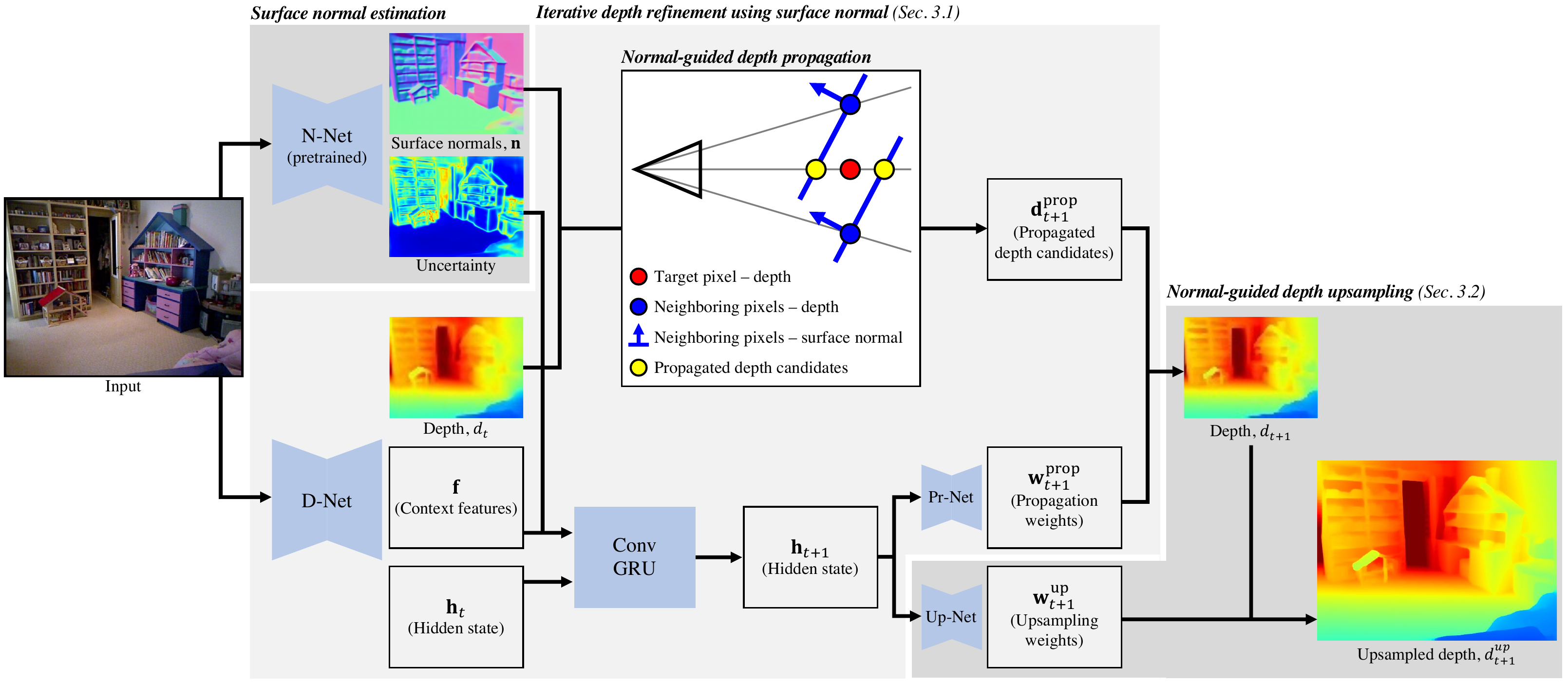}
\end{center}
\caption{This figure illustrates the proposed IronDepth pipeline. Given the input image, N-Net estimates the pixel-wise surface normal and its uncertainty. D-Net estimates the initial low-resolution depth-map $d_{t=0}$. It also produces the context features $\textbf{f}$ and hidden state $\textbf{h}_{t=0}$, which are passed through a ConvGRU~\cite{GRU} cell to produce $\textbf{h}_{t+1}$. With $\textbf{n}$ and $d_t$, we can \textit{propagate} the depth of the neighboring pixels to generate a set of depth candidates $\textbf{d}^\text{prop}_{t+1}$. The updated depth-map $d_{t+1}$ is then given as the weighted sum $\sum \textbf{d}^\text{prop}_{t+1} \textbf{w}^\text{prop}_{t+1}$, where $\textbf{w}^\text{prop}_{t+1}$ is estimated from $\textbf{h}_{t+1}$ using a lightweight CNN (Pr-Net). Lastly, we apply the same principle (normal-guided depth propagation $\rightarrow$ weighted sum) for sub-pixel points to obtain the full-resolution output.}
\label{fig:method}
\end{figure}

\section{Method}

The proposed pipeline is illustrated in Fig.~\ref{fig:method}. It takes a single RGB image with known camera intrinsics as input. Firstly, we use an off-the-shelf network~\cite{SNfromRGB_21_BAE} to estimate the pixel-wise surface normal and its uncertainty. Secondly, D-Net estimates an initial low-resolution depth-map, which is refined iteratively using the predicted surface normal as guidance (Sec.~\ref{sec:method2}). Lastly, we propose normal-guided upsampling to recover the full resolution output (Sec.~\ref{sec:method3}).

\subsection{Iterative depth refinement using surface normal}
\label{sec:method2}

While the predicted surface normal cannot give us the metric depth of a pixel, it tells us how the depth should change around each pixel. Our goal is to exploit such geometric constraint to improve the initial, unconstrained depth prediction. Firstly, we estimate an initial depth-map $d_{t=0}$ using a convolutional encoder-decoder (D-Net). Then, for each pixel, the depths of the neighboring pixels are \textit{propagated} towards the central pixel, using the surface normal as guidance. The weighted sum of the propagated depths then gives us the updated depth-map $d_{t+1}$. To ensure computational efficiency, the refinement is performed in a coarse resolution.

\noindent
\textbf{Initial depth prediction.} The initial depth-map, $d_{t=0}$, is estimated with D-Net, a lightweight convolutional encoder-decoder with EfficientNet B5~\cite{efficientnet} backbone. The architecture is same as the one used in~\cite{mono-2020-adabins}, except that we only decode until $H/8 \times W/8$ resolution, where $H$ and $W$ are the input height and width. Using the decoded feature-map as input, three sets of convolutional layers estimate (1) the initial depth-map $d_{t=0}$, (2) context feature $\textbf{f}$ and (3) the initial hidden state $\textbf{h}_{t=0}$, all in $H/8 \times W/8$ resolution.

\noindent
\textbf{Hidden state update.} The hidden state $\mathbf{h}_t$ is updated recurrently using a Convolutional Gated Recurrent Unit~\cite{GRU} (ConvGRU). We use the architecture of~\cite{RAFT}. The input to the ConvGRU cell is the concatenation of the context feature $\mathbf{f}$ and the surface normal confidence $\kappa$. Since higher value of $\kappa$ means that the predicted surface normal has lower uncertainty, the network can learn to propagate from the neighboring pixel with high $\kappa$.

\noindent
\textbf{Recurrent depth refinement.} Consider a pixel $i$ with pixel coordinates $(u_i,v_i)$. Assuming a pinhole camera, its camera-centered coordinates $\mathbf{X}^c_t$ can be given as
\begin{equation}
\label{eqn:cam_coord}
\mathbf{X}^c_t (u_i,v_i) \!=\! 
\begin{bmatrix} 
\frac{u_i - u_0}{\alpha_u} \\ 
\frac{v_i - v_0}{\alpha_v} \\ 
1 \end{bmatrix} 
\cdot
d_t(u_i,v_i)
=
\mathbf{r}(u_i,v_i) 
\cdot
d_t(u_i,v_i)
\;\;\text{,}\;\;
\mathbf{K} =
\begin{bmatrix} \alpha_u&0&u_0 \\ 0&\alpha_v&v_0 \\ 0&0&1 \end{bmatrix},
\end{equation}

\noindent
where $\mathbf{r}(u_i,v_i)$ represents a ray with unit depth, $\mathbf{K}$ is the camera calibration matrix, and $t$ indexes the iteratively updated depth-map. 

Now, consider a local neighborhood of pixel $i$, which can be defined as $\mathcal{N}_i=\{ j \;:\; |u_i-u_j| \leq \beta \; \text{and} \; |v_i-v_j| \leq \beta \}$. We use $\beta=2$ in all experiments (i.e. $5\times 5$ neighborhood) as it led to a good balance between accuracy and computational efficiency. If the pixel $i$ belongs to the same plane as a neighboring pixel $j$ with depth $d_t(u_j,v_j)$ and surface normal $\mathbf{n}(u_j,v_j)$, its depth should be
\begin{equation}
\label{eqn:depth_prop}
d^\text{prop}_{t+1}(u_i,v_i,j) = 
\frac{\mathbf{n}^\intercal (u_j,v_j) \mathbf{r}(u_j,v_j)}{\mathbf{n}^\intercal(u_j,v_j) \mathbf{r}(u_i,v_i)}
d_t(u_j,v_j).
\end{equation}

We call this the normal-guided depth propagation. In order for pixel $i$ to belong to the same plane as its neighboring pixel $j$, its depth should be updated to $d^\text{prop}_{t+1}(u_i,v_i,j)$. The values of $d^\text{prop}_{t+1}(u_i,v_i,j)$, computed for $j \in \mathcal{N}_i$, can be considered as the per-pixel \textit{candidates} for the refined depth-map. The updated depth-map can thus be given as
\begin{equation}
\label{eqn:depth_update}
d_{t+1}(u_i,v_i) = \sum_{j \in \mathcal{N}_i} w^\text{prop}_{t+1}(u_i,v_i,j) \cdot d^\text{prop}_{t+1}(u_i,v_i,j),
\end{equation}

\noindent
where $\mathbf{w}^\text{prop}_{t+1}(u_i,v_i) = \{ w^\text{prop}_{t+1}(u_i,v_i,j) \}$ is estimated from the hidden state $\mathbf{h}_{t+1}$, using a light-weight CNN (see Appendix for the architecture). Eq.~\ref{eqn:depth_update} shows that depth refinement can be formulated as a $K$-class classification, where $K$ is the number of pixels in the neighborhood $\mathcal{N}_i$. Note that the neighborhood $\mathcal{N}_i$ also includes the pixel $i$ itself, in which case $d^\text{prop}_{t+1}(u_i,v_i,i)=d_t(u_i,v_i)$. The network can thus choose \textit{not} to update depth for certain pixels. 

Normal-guided depth propagation (Eq.~\ref{eqn:depth_prop}) is a view-dependent operation, which depends on the pixels coordinates $(u_i,v_i)$ and $(u_j,v_j)$. However, once the depth candidates $d^\text{prop}_{t+1}(u_i,v_i,j)$ are computed, choosing from them requires view-independent inference, making the problem easier for the network to learn.


\subsection{Normal-guided depth upsampling} 
\label{sec:method3}

For computational efficiency, the depth-map is refined in a coarse resolution ($H/8 \times W/8$). After refinement, the depth-map should be upsampled to match the input resolution. However, linearly upsampling the depth-map does not preserve the surface normal (see Fig.~\ref{fig:motivation}). To this end, we introduce normal-guided upsampling. 

\noindent
\textbf{Normal-guided depth upsampling.} For each pixel in high-resolution depth-map, we can propagate the depths of its $3\times 3$ neighbors in the coarse depth-map. Then, a lightweight CNN (i.e. Up-Net in Fig.~\ref{fig:method}) solves 9-class classification of choosing the coarse resolution neighbor to propagate from. The weighted sum of the propagated depth candidates gives us the upsampled depth-map $d^\text{up}_t$. We show in the experiments that using the proposed normal-guided upsampling leads to better surface normal accuracy than using bilinear upsampling.

\noindent
\textbf{Network training.} The initial depth-map $d_0$, estimated by D-Net, is recurrently refined and upsampled for $N_\text{iter}$ times, producing $\{d^\text{up}_t\}$ where $t \in \{0, 1, ..., N_\text{iter} \}$. The loss is computed as a weighted sum of their L1 losses,
\begin{equation}
\label{eqn:depth_loss}
\mathcal{L}^\text{depth}
=
\sum_{i=0}^{N_\text{iter}}
\gamma^{N_\text{iter} - i}
||d^\text{gt} - d^\text{up}_t||_1, 
\end{equation}

\noindent
where $0 < \gamma < 1$ puts a bigger emphasis on the final output. Following~\cite{RAFT}, we set $\gamma=0.8$. $N_\text{iter}$ is set to 3 during training and 20 at test time.

\section{Experimental Setup}

\noindent
\textbf{Datasets.} Our method is trained and tested on NYUv2~\cite{NYUv2}, which consists of RGB-D frames covering 464 indoor scenes. After training, we also evaluate the network on iBims-1~\cite{iBims} (contains 100 RGB-D frames) without fine-tuning to test its generalization ability. The ground truth surface normal for~\cite{iBims} is obtained by running PCA with $7\times 7$ neighborhood.

\noindent
\textbf{Evaluation protocol.} We evaluate the refined depth-map both in terms of depth and normal. Depth accuracy is evaluated using the metrics defined in~\cite{mono-2014-eigen}. We also compute surface normals from the predicted depth-map, by running per-pixel PCA with $7\times 7$ neighborhood. Then, the angular error between the computed surface normal and the ground truth is measured. Following~\cite{SNfromRGB_13_3DP}, we report the mean, median and root-mean-squared error (lower is better). We also report the percentage of pixels with error less than $[11.25^\circ, 22.5^\circ, 30^\circ]$ (higher is better).

\noindent
\textbf{Implementation details.} The proposed pipeline is implemented with PyTorch~\cite{PyTorch}. We use the AdamW optimizer~\cite{AdamW_introduced} and schedule the learning rate using~\cite{1cycle-lr} with $lr_\text{max} =  3.5 \times 10^{-4}$. We train N-Net for 5 epochs with a batch size of 16. The other components are trained for 10 epochs with a batch size of 4. The EfficientNet~\cite{efficientnet} backbone of D-Net is fixed with the weights from~\cite{mono-2020-adabins}.

\begin{table}[t]
\small
\setlength\tabcolsep{1.0pt}
\begin{center}
\begin{tabular}{l|ccc|ccc|ccc|ccc}
\toprule
\multirow{2}{4em}{Method} 
& \multicolumn{3}{c|}{Depth error} 
& \multicolumn{3}{c|}{Depth accuracy}
& \multicolumn{3}{c|}{Normal error}
& \multicolumn{3}{c}{Normal accuracy}\\
\cline{2-13} 
& abs rel & rmse & $\log_{10}$ & $\delta_1$ & $\delta_2$ & $\delta_3$ & mean & median & rmse & $11.25^{\circ}$ & $22.5^{\circ}$ & $30^{\circ}$ \\
\midrule
GeoNet~\cite{SNfromRGB_18_GeoNet} & 
0.142 & 0.499 & 0.062 & 0.801 & 0.963 & 0.992 &
41.5 & 35.5 & 50.2 & 11.7 & 30.5 & 42.2 \\
DORN~\cite{mono-2018-DORN} & 
0.106 & 0.397 & 0.046 & 0.877 & 0.970 & 0.990 &
44.7 & 39.3 & 53.3 & 9.2 & 26.7 & 38.0 \\
VNL~\cite{mono-2019-VNL} & 
\textbf{0.100} & 0.368 & \textbf{0.043} & 0.895 & 0.980 & 0.996 &
26.8 & 17.0 & 37.9 & 36.3 & 59.4 & 68.6 \\
BTS~\cite{mono-2019-BTS} & 
0.110 & 0.392 & 0.047 & 0.886 & 0.978 & 0.994 &
32.4 & 24.7 & 42.1 & 22.7 & 46.1 & 58.3 \\
AdaBins~\cite{mono-2020-adabins} & 
0.103 & 0.364 & 0.044 & 0.902 & 0.983 & \textbf{0.997} &
28.8 & 20.7 & 38.6 & 28.3 & 53.2 & 64.7 \\
TransDepth~\cite{mono-2021-transdepth} & 
0.106 & 0.365 & 0.045 & 0.900 & 0.983 & 0.996 &
30.0 & 22.4 & 39.7 & 25.6 & 50.2 & 62.4 \\
\hline
Ours &
0.101 & \textbf{0.352} & \textbf{0.043} & \textbf{0.910} & \textbf{0.985} & \textbf{0.997} & 
\textbf{20.8} & \textbf{11.3} & \textbf{31.9} & 
\textbf{49.7} & \textbf{70.5} & \textbf{77.9} \\
\bottomrule
\end{tabular}
\end{center}
\caption{Quantitative evaluation on NYUv2~\cite{NYUv2}. We show state-of-the-art performance both in terms of depth and normal.}
\label{table:depth_NYUv2}
\end{table}

\section{Experiments}

In Sec.~\ref{sec:exp2}, we evaluate our method on NYUv2~\cite{NYUv2} and iBims-1~\cite{iBims}. We make quantitative and qualitative comparison against the state-of-the-art methods and run ablation study experiments. In Sec.~\ref{sec:exp3}, we explore the usefulness of the proposed normal-guided depth propagation.

\begin{figure}[t]
\begin{center}
\includegraphics[width=1.0\linewidth]{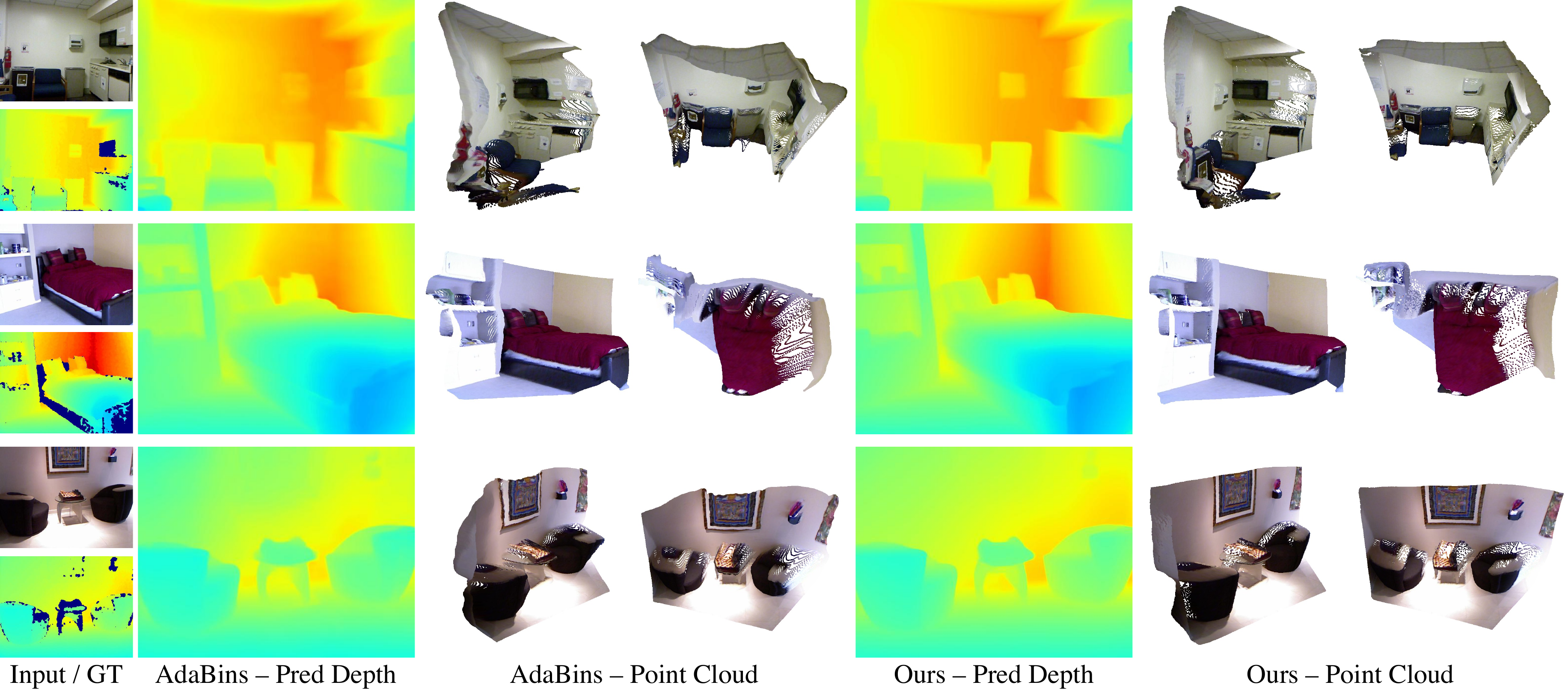}
\end{center}
\caption{This figure provides a qualitative comparison between AdaBins~\cite{mono-2020-adabins} and our method. While the predicted depth-maps look similar, the point cloud comparison shows that our method is better at capturing the orientation of the surfaces.}
\label{fig:benchmark}
\end{figure}

\begin{figure}[t]
\begin{center}
\includegraphics[width=1.0\linewidth]{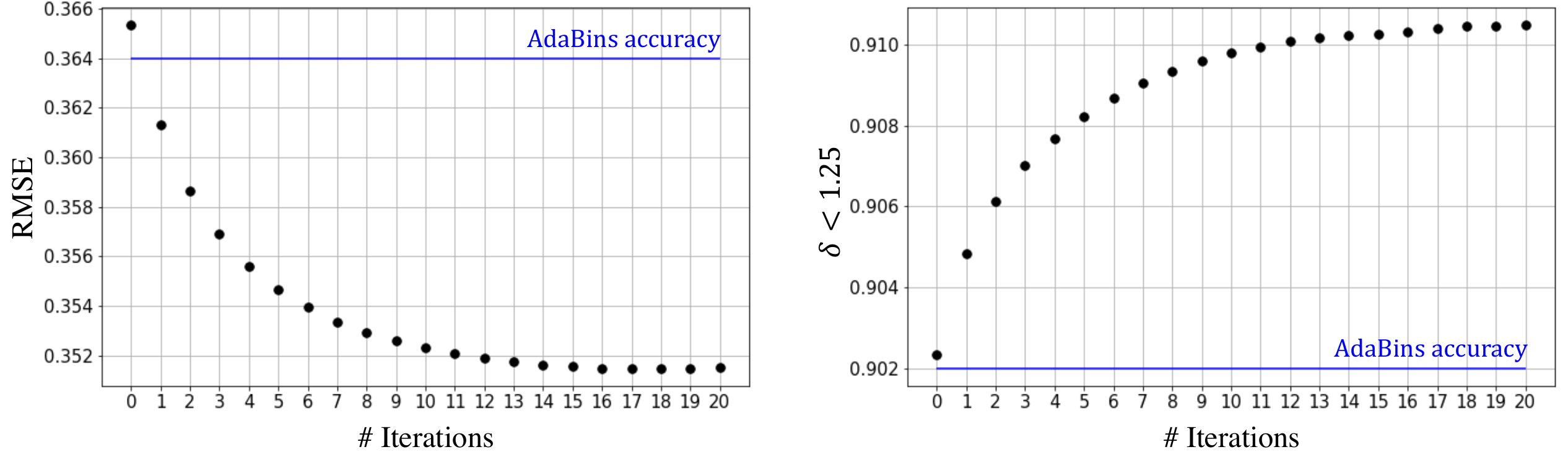}
\end{center}
\caption{This figure shows how the accuracy on NYUv2~\cite{NYUv2} improves during the normal-guided iterative refinement. The accuracy converges after about 10 iterations.}
\label{fig:convergence}
\end{figure}

\begin{table}[t]
\small
\setlength\tabcolsep{1.0pt}
\begin{center}
\begin{tabular}{l|ccc|ccc|ccc|ccc|cc}
\toprule
\multirow{2}{4em}{Method} 
& \multicolumn{3}{c|}{Depth error} 
& \multicolumn{3}{c|}{Depth accuracy}
& \multicolumn{3}{c|}{Normal error} 
& \multicolumn{3}{c|}{Normal accuracy}
& \multicolumn{2}{c}{Planarity} \\
\cline{2-15} 
& rel & rmse & $\log_{10}$ & $\delta_1$ & $\delta_2$ & $\delta_3$ 
& mean & median & rmse & $11.25^{\circ}$ & $22.5^{\circ}$ & $30^{\circ}$
& $\epsilon^\text{plan}$ & $\epsilon^\text{orie}$ \\
\midrule
SharpNet~\cite{mono-2019-SharpNet}
& 0.26 & 1.07 & 0.11 & 0.59 & 0.84 & 0.94 
& - & - & - & - & - & -
& 9.95 & 25.67 \\
VNL~\cite{mono-2019-VNL}
& 0.24 & 1.07 & 0.11 & 0.55 & 0.85 & 0.94 
& 39.8 & 30.4 & 51.0 & 17.9 & 38.6 & 49.4
& 6.49 & 18.72 \\
BTS~\cite{mono-2019-BTS}
& 0.24 & 1.08 & 0.12 & 0.53 & 0.84 & 0.94 
& 44.0 & 37.8 & 53.5 & 13.0 & 29.5 & 40.0
& 7.25 & 20.52 \\
DAV~\cite{mono-2020-dav} 
& 0.24 & 1.06 & \textbf{0.10} & \textbf{0.59} & 0.84 & 0.94
& - & - & - & - & - & - 
& 7.21 & 18.45 \\
AdaBins~\cite{mono-2020-adabins}
& 0.22 & 1.06 & 0.11 & 0.55 & 0.86 & \textbf{0.95}
& 37.1 & 29.6 & 46.9 & 18.0 & 38.7 & 50.6 
& 6.25 & 17.51 \\
\hline
Ours 
& \textbf{0.21} & \textbf{1.03} & 0.11 
& \textbf{0.59} & \textbf{0.87} & \textbf{0.95} 
& \textbf{25.3} & \textbf{14.2} & \textbf{37.4} 
& \textbf{43.1} & \textbf{63.9} & \textbf{71.6} 
& \textbf{3.29} & \textbf{8.48} \\
\bottomrule
\end{tabular}
\end{center}
\caption{Cross-dataset evaluation on iBims-1~\cite{iBims}. Our method shows significantly higher surface normal accuracy. We also outperform other methods in terms of planarity (quantifies how planar the prediction is for walls, table surfaces and floors).}
\label{table:depth_ibims}
\end{table}


\subsection{Main results}
\label{sec:exp2}


\noindent
\textbf{NYUv2.} Tab.~\ref{table:depth_NYUv2} shows that our method achieves state-of-the-art performance on NYUv2~\cite{NYUv2}. While the differences in the depth metrics are small, the surface normals computed from our depth-maps are significantly more accurate than those obtained by the other methods. For example, the mean angular error ($20.8^\circ$) is 22.4\% smaller than the second best method ($26.8^\circ$ achieved by~\cite{mono-2019-VNL}). Fig.~\ref{fig:benchmark} provides a qualitative comparison against~\cite{mono-2020-adabins}. The point cloud comparison shows that our method faithfully captures the surface layout of the scene. Fig.~\ref{fig:convergence} shows how the accuracy improves during the iterative refinement. Before refinement, the accuracy is similar to that of~\cite{mono-2020-adabins}. The accuracy improves quickly in the first few iterations and converges after about 10 iterations.

\noindent
\textbf{iBims-1.} Tab.~\ref{table:depth_ibims} evaluates the generalization ability on iBims-1~\cite{iBims}. Similar to the results on NYUv2, we show a small improvement in depth accuracy, but a large improvement in surface normal accuracy. We also report $\epsilon^\text{plan}$ and $\epsilon^\text{orie}$ (defined in~\cite{iBims}), which quantify the planarity of the pixels belonging to walls, table surfaces and floors. We achieve 47.4\% reduction in $\epsilon^\text{plan}$ and 51.6\% reduction in $\epsilon^\text{orie}$, compared to AdaBins~\cite{mono-2020-adabins}.


\begin{figure}[t]
\begin{center}
\includegraphics[width=1.0\linewidth]{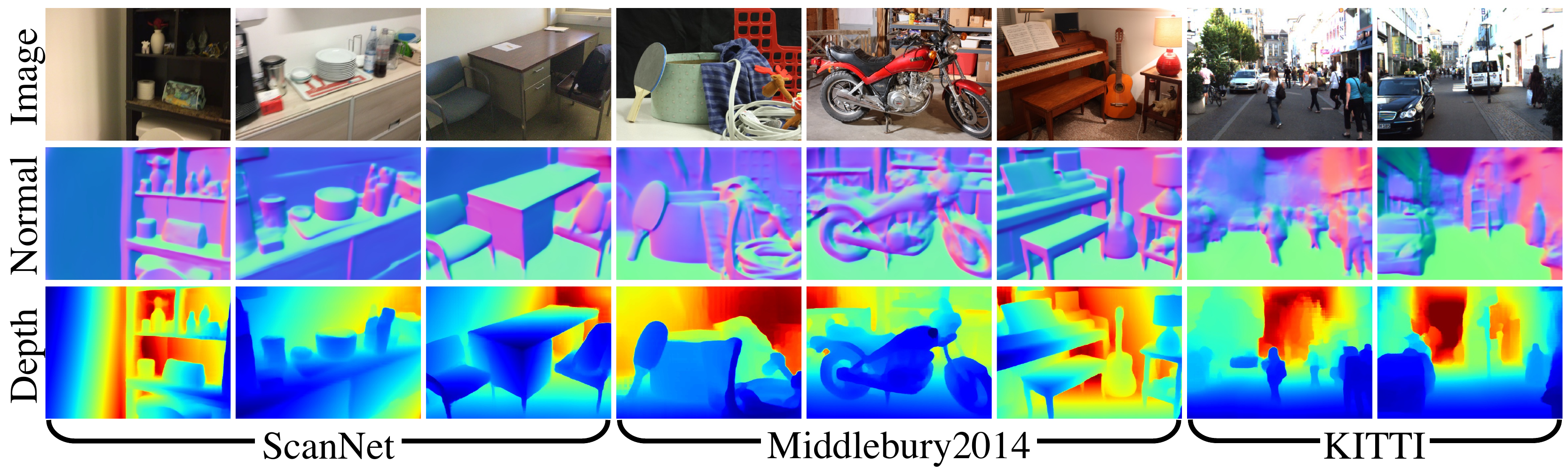}
\end{center}
\caption{This figure shows predictions made by our method trained on NYUv2~\cite{NYUv2}. The network generalizes well to ScanNet~\cite{ScanNet}, Middlebury~\cite{Middlebury} and KITTI~\cite{KITTI}.}
\label{fig:generalize}
\end{figure}

\noindent
\textbf{Generalization.} In Fig.~\ref{fig:generalize}, we further demonstrate the generalization ability of our method. As highlighted in~\cite{SNfromRGB_21_BAE}, surface normal estimation networks generalize well across different datasets as they rely on low-level cues (e.g., texture gradients, shading). Since IronDepth uses the predicted surface normal to refine the initial depth-map, it can generalize well even when the domain gap is large (e.g., train on indoor scenes $\rightarrow$ test on outdoor scenes).

\begin{table}[t]
\small
\setlength\tabcolsep{1.0pt}
\begin{center}
\begin{tabular}{cc|ccc|ccc|ccc|ccc}
\toprule
Iterative & Upsample
& \multicolumn{3}{c|}{Depth error} 
& \multicolumn{3}{c|}{Depth accuracy}
& \multicolumn{3}{c|}{Normal error}
& \multicolumn{3}{c}{Normal accuracy}\\
\cline{3-14} 
Refinement & Method
& abs rel & rmse & $\log_{10}$ & $\delta_1$ & $\delta_2$ & $\delta_3$ & mean & median & rmse & $11.25^{\circ}$ & $22.5^{\circ}$ & $30^{\circ}$ \\
\midrule
$\times$ & \multirow{2}{*}{Nearest} &
0.107 & 0.375 & 0.045 & 0.896 & 0.982 & 0.996 &
39.1 & 31.6 & 47.8 & 11.6 & 35.7 & 47.8 \\
$\checkmark$ &  &
0.103 & 0.361 & 0.044 & 0.904 & 0.984 & \textbf{0.997} &
33.7 & 23.6 & 43.1 & 17.5 & 47.9 & 59.4 \\
\hline
$\times$ & \multirow{2}{*}{Bilinear} &
0.106 & 0.370 & 0.045 & 0.898 & 0.983 & 0.996 &
31.4 & 22.9 & 41.6 & 25.1 & 49.3 & 60.9 \\
$\checkmark$ &  &
0.103 & 0.356 & 0.044 & 0.906 & \textbf{0.985} & \textbf{0.997} &
22.7 & 12.2 & 34.6 & 47.6 & 67.6 & 75.0 \\
\hline
$\times$ & Normal &
0.105 & 0.369 & 0.045 & 0.897 & 0.982 & 0.996 & 
28.6 & 20.0 & 39.0 & 30.8 & 54.4 & 65.0 \\
$\checkmark$ & -guided &
\textbf{0.101} & \textbf{0.352} & \textbf{0.043} & \textbf{0.910} & \textbf{0.985} & \textbf{0.997} &
\textbf{20.8} & \textbf{11.3} & \textbf{31.9} & \textbf{49.7} & \textbf{70.5} & \textbf{77.9} \\
\bottomrule
\end{tabular}
\end{center}
\caption{Ablation study experiments. We train the pipeline with and without the iterative refinement, and also try different methods of depth upsampling.}
\label{table:ablation1}
\end{table}

\noindent
\textbf{Ablation study.} Tab.~\ref{table:ablation1} provides the results of the ablation study experiments. Iterative depth refinement with normal-guided depth propagation (Sec.~\ref{sec:method2}) significantly improves the accuracy, both in terms of depth and normal. Compared to bilinear upsampling, the proposed normal-guided upsampling (Sec.~\ref{sec:method3}) leads to better surface normal accuracy. 



\noindent
\textbf{Inference speed.} The inference time of the full pipeline is 66.26 ms, when measured on a single 2080Ti GPU. The proposed normal-guided depth refinement only takes 0.57ms per iteration. This is because the refinement is performed in a coarse resolution ($H/8 \times W/8$).

\subsection{Applications}
\label{sec:exp3}

Lastly, we discuss the possible applications of the proposed framework.

\begin{table}[t]
\small
\setlength\tabcolsep{1.0pt}
\begin{center}
\begin{tabular}{l|ccc|ccc|ccc|ccc}
\toprule
\multirow{2}{4em}{Method} 
& \multicolumn{3}{c|}{Depth error}
& \multicolumn{3}{c|}{Depth accuracy}
& \multicolumn{3}{c|}{Normal error}
& \multicolumn{3}{c}{Normal accuracy}\\
\cline{2-13} 
& abs rel & rmse & $\log_{10}$ & $\delta_1$ & $\delta_2$ & $\delta_3$ 
& mean & median & rmse & $11.25^{\circ}$ & $22.5^{\circ}$ & $30^{\circ}$ \\
\midrule
DORN~\cite{mono-2018-DORN} & 
0.106 & 0.397 & 0.046 & 0.877 & 0.970 & 0.990 &
44.7 & 39.3 & 53.3 & 9.2 & 26.7 & 38.0 \\
DORN + Ours &
\textbf{0.099} & \textbf{0.359} & \textbf{0.042} & 
\textbf{0.898} & \textbf{0.978} & \textbf{0.993} & 
\textbf{21.3} & \textbf{11.8} & \textbf{32.5} & 
\textbf{48.5} & \textbf{69.6} & \textbf{77.1} \\
\hline
VNL~\cite{mono-2019-VNL} & 
0.100 & 0.368 & 0.043 & 0.895 & 0.980 & 0.996 &
26.8 & 17.0 & 37.9 & 36.3 & 59.4 & 68.6 \\
VNL + Ours &
\textbf{0.097} & \textbf{0.353} & \textbf{0.042} & 
\textbf{0.902} & \textbf{0.983} & \textbf{0.996} & 
\textbf{20.5} & \textbf{11.0} & \textbf{31.7} & 
\textbf{50.6} & \textbf{71.0} & \textbf{78.2} \\
\hline
BTS~\cite{mono-2019-BTS} & 
0.110 & 0.392 & 0.047 & 0.886 & 0.978 & 0.994 &
32.4 & 24.7 & 42.1 & 22.7 & 46.1 & 58.3 \\
BTS + Ours &
\textbf{0.104} & \textbf{0.368} & \textbf{0.044} & 
\textbf{0.899} & \textbf{0.981} & \textbf{0.995} & 
\textbf{21.0} & \textbf{11.5} & \textbf{32.1} & 
\textbf{49.4} & \textbf{70.2} & \textbf{77.6} \\
\hline
AdaBins~\cite{mono-2020-adabins} & 
0.103 & 0.364 & 0.044 & 0.902 & 0.983 & 0.997 &
28.8 & 20.7 & 38.6 & 28.3 & 53.2 & 64.7 \\
AdaBins + Ours & 
\textbf{0.100} & \textbf{0.351} & \textbf{0.042} & 
\textbf{0.911} & \textbf{0.985} & \textbf{0.997} & 
\textbf{20.7} & \textbf{11.3} & \textbf{31.8} & 
\textbf{49.9} & \textbf{70.6} & \textbf{78.0} \\
\hline
TransDepth~\cite{mono-2021-transdepth} & 
0.106 & 0.365 & 0.045 & 0.900 & 0.983 & 0.996 &
30.0 & 22.4 & 39.7 & 25.6 & 50.2 & 62.4 \\
TransDepth + Ours &
\textbf{0.103} & \textbf{0.352} & \textbf{0.043} & 
\textbf{0.906} & \textbf{0.984} & \textbf{0.997} & 
\textbf{20.6} & \textbf{11.1} & \textbf{31.7} & 
\textbf{50.3} & \textbf{70.9} & \textbf{78.3} \\
\bottomrule
\end{tabular}
\end{center}
\caption{Normal-guided depth refinement applied to the existing depth estimation methods. The accuracy is improved across all metrics.}
\label{table:sota_plus_ours}
\end{table}


\noindent
\textbf{Application to existing depth estimation methods.} The normal-guided depth refinement can be applied to the predictions made by the existing depth estimation methods. Specifically, we can replace the $d_0$ estimated by D-Net with the predictions made by other methods (the network is not fine-tuned for each method). Tab.~\ref{table:sota_plus_ours} shows that the accuracy is improved across all metrics. Significant improvement in the surface normal accuracy suggests that our framework can be used as a post-processing tool to improve the surface normal accuracy of the existing monocular depth estimation methods. This also suggests that replacing our D-Net (i.e. lightweight convolutional encoder-decoder) with a more sophisticated architecture can further improve the accuracy.


\begin{table}[t]
\small
\setlength\tabcolsep{1.0pt}
\begin{center}
\begin{tabular}{l|ccc|ccc|ccc|ccc}
\toprule
\multirow{2}{*}{\# Measurements} 
& \multicolumn{6}{c|}{Depth metrics (w/o scale-match)}
& \multicolumn{6}{c}{Depth metrics (w/ scale-match)}
\\
\cline{2-13} 
& abs rel & rmse & $\log_{10}$ & $\delta_1$ & $\delta_2$ & $\delta_3$
& abs rel & rmse & $\log_{10}$ & $\delta_1$ & $\delta_2$ & $\delta_3$ \\
\midrule
0
& 0.101 & 0.352 & 0.043 & 0.910 & 0.985 & 0.997
& 0.101 & 0.352 & 0.043 & 0.910 & 0.985 & 0.997 \\
\hline
10 
& 0.097 & 0.341 & 0.041 & 0.917 & 0.986 & 0.997
& 0.076 & 0.300 & 0.033 & 0.944 & 0.991 & 0.998 \\
50
& 0.084 & 0.304 & 0.035 & 0.938 & 0.990 & 0.998
& 0.063 & 0.260 & 0.027 & 0.962 & 0.994 & 0.999 \\
100
& 0.070 & 0.266 & 0.030 & 0.957 & 0.993 & 0.999
& 0.053 & 0.231 & 0.023 & 0.972 & 0.995 & 0.999 \\
200
& 0.051 & 0.212 & 0.021 & 0.976 & 0.996 & 0.999
& 0.041 & 0.191 & 0.018 & 0.983 & 0.997 & 0.999 \\
\bottomrule
\end{tabular}
\end{center}
\caption{We provide the ground truth for a small number of pixels and fix their values. The depths of those pixels are propagated to the neighboring pixels, improving the overall accuracy. We can also multiply the initial prediction by a factor that minimizes the error for the anchor points (before performing the refinement). Columns 8-13 show that applying such scale-matching leads to a bigger improvement.}
\label{table:anchor_points}
\end{table}

\noindent
\textbf{Application to depth completion.} Suppose that a network is trained to estimate depth from a single RGB image. If a new piece of information (e.g., sparse depth measurement from a LiDAR sensor) is available at test time, the network should be able to \textit{adapt} to that information and the prediction should be more accurate. However, such ability to adapt is not possessed by most depth estimation methods. Since we refine the depth map by \textit{propagating} information between the pixels, we can seamlessly apply our method to a scenario where sparse depth measurements are available (i.e. depth completion setup). Given a sparse depth measurement, we can add \textit{anchor points} by fixing the depth for the pixels with measurement. We simulate this by providing the ground truth for a small number of pixels. Tab.~\ref{table:anchor_points} shows how the accuracy can be improved by adding such anchor points. The information provided for the anchor points (i.e. the measured depth) can be propagated to the neighboring pixels, making the overall prediction more accurate.  




\section{Conclusions}

In this work, we proposed IronDepth, a novel framework that uses surface normal and its uncertainty to recurrently refine the predicted depth-map. We used normal-guided depth propagation to formulate depth refinement as classification of choosing the neighboring pixel to propagate from. Our method achieves state-of-the-art performance on NYUv2~\cite{NYUv2} and iBims-1~\cite{iBims}, both in terms of depth and surface normal. Point cloud comparison shows that our method is better at capturing the surface layout of the scene. The proposed framework can also be used as a post-processing tool for the existing depth estimation methods, or to propagate a sparse depth measurement to improve the overall accuracy.

\noindent 
\textbf{Acknowledgement.} This research was sponsored by Toshiba Europe's Cambridge Research Laboratory.

\appendix

\section{Network Architecture}

Tab. \ref{table:archi_dnet} shows the architecture of D-Net, which estimates the initial depth-map $d_{t=0}$, context feature $\textbf{f}$ and the initial hidden state $\mathbf{h}_{t=0}$. Tab. \ref{table:archi_prnet} shows the architecture of Pr-Net, which estimates the propagation weights $\mathbf{w}^\text{prop}_t$ for each iteration. Tab. \ref{table:archi_upnet} shows the architecture of Up-Net, which estimates the upsampling weights $\mathbf{w}^\text{up}_t$.

\begin{table}[h]
\footnotesize
\setlength{\tabcolsep}{2.6pt}
\begin{center}
\begin{tabular}{c|c|c|c}
\hline
\textbf{Input} & \textbf{Layer} & \textbf{Output} & \textbf{Output Dimension} \\
\hline
\textit{image} & - & - & $H \times W \times 3$ \\
\hline
\multicolumn{4}{c}{\textbf{Encoder}} \\
\hline
\multirow{3}{*}{\textit{image}}
& \multirow{3}{*}{EfficientNet B5}
& \textit{$F_8$} & $H/8 \times W/8 \times 64$ \\
& & \textit{$F_{16}$} & $H/16 \times W/16 \times 176$ \\
& & \textit{$F_{32}$} & $H/32 \times W/32 \times 2048$ \\
\hline
\multicolumn{4}{c}{\textbf{Decoder}} \\
\hline
\textit{$F_{32}$} & Conv2D(ks=1, $C_\text{out}$=2048, padding=0) 
& $x_0$
& $H/32 \times W/32 \times 2048$ \\
\hline
$\text{up}(x_0) + F_{16}$
& 
$\left( 
\begin{matrix} 
\text{Conv2D(ks=3, $C_\text{out}$=1024, padding=1)}, \\
\text{GroupNorm}(n_\text{groups}=8), \\
\text{LeakyReLU()}
\end{matrix}
\right)
\times 2$ 
& $x_1$ & $H/16 \times W/16 \times 1024$ \\
\hline
$\text{up}(x_1) + F_{8}$
& 
$\left( 
\begin{matrix} 
\text{Conv2D(ks=3, $C_\text{out}$=512, padding=1)}, \\
\text{GroupNorm}(n_\text{groups}=8), \\
\text{LeakyReLU()}
\end{matrix}
\right)
\times 2$ 
& $x_2$ & $H/8 \times W/8 \times 512$ \\
\hline
\multicolumn{4}{c}{\textbf{Prediction Heads}} \\
\hline
$x_2$ & 
\makecell{
Conv2D(ks=3, $C_\text{out}$=128, padding=1), ReLU(), \\
Conv2D(ks=1, $C_\text{out}$=128, padding=0), ReLU(), \\
Conv2D(ks=1, $C_\text{out}$=1, padding=0) \\
}
& $d_{t=0}$ & $H/8 \times W/8 \times 1$ \\
\hline
$x_2$ & 
\makecell{
Conv2D(ks=3, $C_\text{out}$=128, padding=1), ReLU(), \\
Conv2D(ks=1, $C_\text{out}$=128, padding=0), ReLU(), \\
Conv2D(ks=1, $C_\text{out}$=64, padding=0) \\
}
& $\mathbf{f}$ & $H/8 \times W/8 \times 64$ \\
\hline
$x_2$ & 
\makecell{
Conv2D(ks=3, $C_\text{out}$=128, padding=1), ReLU(), \\
Conv2D(ks=1, $C_\text{out}$=128, padding=0), ReLU(), \\
Conv2D(ks=1, $C_\text{out}$=64, padding=0) \\
}
& $\mathbf{h}_{t=0}$ & $H/8 \times W/8 \times 64$ \\
\hline
\end{tabular}
\end{center}
\caption{D-Net architecture. In each convolutional layer, "ks" means the kernel size and $C_\text{out}$ is the number of output channels. $F_N$ represents the feature-map of resolution $H/N \times W/N$. $X + Y$ means that the two tensors are concatenated, and $\text{up}(\cdot)$ is bilinear upsampling.}
\label{table:archi_dnet}
\end{table}

\begin{table}[h]
\footnotesize
\setlength{\tabcolsep}{2.6pt}
\begin{center}
\begin{tabular}{c|c|c|c}
\hline
\textbf{Input} & \textbf{Layer} & \textbf{Output} & \textbf{Output Dimension} \\
\hline
$\mathbf{h}_t$
& 
\makecell{
Conv2D(ks=3, $C_\text{out}=128$, padding=1), ReLU(), \\
Conv2D(ks=1, $C_\text{out}=128$, padding=0), ReLU(), \\
Conv2D(ks=1, $C_\text{out}=5\times 5$, padding=0) \\
}
& $\mathbf{w}^\text{prop}_t$ & $H/8 \times W/8 \times (5\times 5)$ \\
\hline
\end{tabular}
\end{center}
\caption{Architecture of Pr-Net. Pr-Net estimates the propagation weights $\mathbf{w}^\text{prop}_t$.}
\label{table:archi_prnet}
\end{table}

\begin{table}[h]
\footnotesize
\setlength{\tabcolsep}{2.6pt}
\begin{center}
\begin{tabular}{c|c|c|c}
\hline
\textbf{Input} & \textbf{Layer} & \textbf{Output} & \textbf{Output Dimension} \\
\hline
$\mathbf{h}_t$
& 
\makecell{
Conv2D(ks=3, $C_\text{out}$=128, padding=1), ReLU(), \\
Conv2D(ks=1, $C_\text{out}$=128, padding=0), ReLU(), \\
Conv2D(ks=1, $C_\text{out}$=$8\times 8\times 9$, padding=0) \\
}
& $\mathbf{w}^\text{up}_t$ & $H/8 \times W/8 \times (8\times8\times9)$ \\
\hline
\end{tabular}
\end{center}
\caption{Architecture of Up-Net. Up-Net estimates the upsampling weights $\mathbf{w}^\text{up}_t$.}
\label{table:archi_upnet}
\end{table}

\bibliography{main}
\end{document}


\maketitle








\section{Demo Video}

In the attached video, we demonstrate how the predicted 3D point cloud gets updated during the proposed normal-guided iterative depth refinement.